\pdfoutput=1

\documentclass[11pt]{article}

\usepackage[]{EMNLP2022}

\usepackage{times}
\usepackage{latexsym}

\usepackage[T1]{fontenc}

\usepackage[utf8]{inputenc}

\usepackage{microtype}

\usepackage{inconsolata}

\usepackage{listings}
\usepackage{placeins} 

\usepackage{color}
\definecolor{ForestGreen}{RGB}{34,139,34}
\definecolor{Amber}{RGB}{255,191,0}

\usepackage{hyperref}

\usepackage[inline]{enumitem}
\usepackage{scrextend}
\usepackage{array}
\usepackage{arydshln}
\usepackage{graphicx}
\usepackage{hhline}

\usepackage{enumitem}

\usepackage{censor}
\usepackage{gb4e}
\noautomath

\usepackage{amsmath}

\usepackage{multirow}
\usepackage{makecell}

\usepackage{arydshln}
\usepackage{array}
\newcolumntype{L}[1]{>{\raggedright\let\newline\\\arraybackslash\hspace{0pt}}m{#1}}
\newcolumntype{C}[1]{>{\centering\let\newline\\\arraybackslash\hspace{0pt}}m{#1}}
\newcolumntype{R}[1]{>{\raggedleft\let\newline\\\arraybackslash\hspace{0pt}}m{#1}}

\usepackage{scrextend}
\newcommand\blfootnote[1]{%
  \begingroup
  \renewcommand\thefootnote{}\footnote{#1}%
  \addtocounter{footnote}{-1}%
  \endgroup
}

\title{Effective Cross-Task Transfer Learning for \\Explainable Natural Language Inference with \emph{T5}}

\author{
Irina Bigoulaeva\textsuperscript{1*},
Rachneet Sachdeva\textsuperscript{1*},
Harish Tayyar Madabushi\textsuperscript{2*},
\\
\textbf{Aline Villavicencio\textsuperscript{3}} \and
\textbf{Iryna Gurevych\textsuperscript{1}}
\\[0.3cm]
\textsuperscript{1} Ubiquitous Knowledge Processing (UKP) Lab, Technische Universität Darmstadt\\
\textsuperscript{2} Department of Computer Science, The University of Bath \\
\textsuperscript{3} Department of Computer Science, The University of Sheffield \\
\texttt{\small www.ukp.tu-darmstadt.de} \\[-0.1mm]
\texttt{\small htm43@bath.ac.uk, a.villavicencio@sheffield.ac.uk}\\
\\
}

\begin{document}
\maketitle
\blfootnote{\hspace*{-0.112cm}\textsuperscript{*}Equal Contribution}
\begin{abstract}
We compare sequential fine-tuning with a model for multi-task learning in the context where we are interested in boosting performance on two tasks, one of which depends on the other. We test these models on the \emph{FigLang2022} shared task which requires participants to predict language inference labels on figurative language along with corresponding textual explanations of the inference predictions. Our results show that while sequential multi-task learning can be tuned to be good at the first of two target tasks, it performs less well on the second and additionally struggles with overfitting. Our findings show that simple sequential fine-tuning of text-to-text models is an extraordinarily powerful method for cross-task knowledge transfer while simultaneously predicting multiple interdependent targets. So much so, that our best model achieved the (tied) \emph{highest score} on the task\footnote{To ensure reproducibility and to enable other researchers to build upon our work, we make our code and models freely available at \url{https://github.com/Rachneet/cross-task-figurative-explanations}}.
\end{abstract}

\section{Introduction and Motivation}
\label{section:intro}

The transfer of information between \emph{supervised learning objectives} can be achieved in Pre-trained Language Models (PLMs) using either multi-task learning (MTL)~\cite{caruana1997multitask} or sequential fine-tuning (SFT)~\cite{DBLP:journals/corr/abs-1811-01088}. MTL involves simultaneously training a model on multiple learning objectives using a weighted sum of their loss, while SFT involves sequentially training on a set of related tasks. Recent work has extended the SFT approach by converting all NLP problems into text-to-text (i.e., sequence-to-sequence where both input and output sequences are natural text) problems~\cite{DBLP:journals/corr/abs-1910-10683}. The resultant model -- \emph{T5} -- has achieved state-of-the-art results on a variety of tasks such as question answering, sentiment analysis, and, most relevant to this work, Natural Language Inference (NLI). 

In this work, we focus our efforts on the transfer of information from multiple related tasks for improved performance on a different set of tasks. In addition, we compare the effectiveness of SFT with that of MTL in a context where one of the target tasks is dependent on the other. Given the dependence of one of the target tasks on the other, we implement an end-to-end multi-task learning model to perform each of the tasks sequentially: an architecture referred to as a \emph{hierarchical feature pipeline} based MTL architecture (\emph{HiFeatMTL}, for short)~\cite{DBLP:journals/corr/abs-2109-09138}. While HiFeatMTL has been previously used in different contexts (see Section \ref{relatedworks}), it has, to the best of our knowledge, \emph{not} been used with, or compared to, text-to-text models. This is of particular importance as such models are known to enable transfer learning~\cite{DBLP:journals/corr/abs-1910-10683} and it is crucial to determine if traditional MTL methods can boost cross-task knowledge transfer in such models.

Specifically we participate in the FigLang2022 Shared Task\footnote{https://figlang2022sharedtask.github.io/}, which extends NLI to include a figurative-language hypothesis and additionally requires participants to output a textual explanation (also see Section \ref{section:thetask}). FigLang2022 is ideally suited for the exploration of knowledge transfer, as PLMs have been shown to struggle with figurative language and so any gains achieved are a result of knowledge transfer. For example, \newcite{liu2022testing} show that in the zero- and few-shot settings, PLMs 
perform significantly worse than humans. This is especially the case with idioms~\cite{yu-ettinger-2020-assessing,tayyar-madabushi-etal-2021-astitchinlanguagemodels-dataset}, on which T5 does particularly poorly (see Section \ref{oursetup}). Additionally, FigLang2022's emphasis on explanations of the predicted labels provides us with the opportunity to test cross-task knowledge transfer in a setting where one target task depends on the other (HiFeatMTL) -- this is especially so given the evaluation methods used (detailed in Section \ref{section:thetask}).

We evaluate the effectiveness of boosting performance on the target tasks through the transfer of information from two related tasks: a) eSNLI, which is a dataset consisting of explanations associated with NLI labels, and b) IMPLI, which is an NLI dataset (without explanations) that contains figurative language. More concretely, we set out to answer the following research questions: 

\begin{enumerate}[topsep=0.2pt]
\itemsep-0.1em 
    \item Can distinct task-specific knowledge be transferred from separate tasks so as to improve performance on a target task? Concretely, can we transfer explanations of literal language from eSNLI and figurative NLI without explanations from IMPLI?
    \item Which of the two knowledge transfer techniques (SFT or HiFeatMTL) is more effective in the text-to-text context?
\end{enumerate}

\section{The FigLang2022 Shared Task}
\label{section:thetask}
FigLang2022 is a variation of the NLI task which requires the generation of a textual explanation for the NLI prediction. Additionally, the hypothesis is a sentence that employs one of four kinds of figurative expressions: \textit{sarcasm, simile, idiom, }or \textit{metaphor}. Additionally, a hypothesis can be a \textit{creative paraphrase}, which rewords the premise using 
more expressive, literal terminology. Table \ref{table:td-examples} shows examples from the task dataset.

\begin{table}[ht]
{\footnotesize
\centering
\resizebox{0.45\textwidth}{!}{
\begin{tabular}{ll}
\hline
\textbf{Entailment} \\\hline
Premise & I respectfully disagree.\\
Hypothesis & I beg to differ.  \textit{(Idiom)} \\
Explanation &  	
To beg to differ is to disagree \\
& with someone, and in this \\
& sentence the speaker is \\
& respectfully disagreeing. \\
\hline
\textbf{Contradiction} \\
\hline
Premise & She was calm. \\
Hypothesis & She was like a kitten in a den \\ 
& of coyotes. \textit{(Simile)} \\
Explanation &  	
A kitten in a den of coyotes \\
& would be scared and not calm. \\
\hline
\end{tabular}
}
\caption{An entailment and contradiction pair from the FigLang2022 dataset.}
\label{table:td-examples}
}
\end{table}

FigLang2022 takes into consideration the quality of the generated explanation when assessing the model's performance by use of an \textit{explanation score}, which is the average between BERTScore and BLEURT and ranges between 0 and 100. The task leaderboard is based on NLI label accuracy at an explanation score threshold of 60, although the NLI label accuracy is reported at three thresholds of the explanation score (i.e. 0, 50, and 60) so as to provide a glimpse of how the model's NLI and explanation abilities are influenced by each other.


\section{Related Work}
\label{relatedworks}
NLI is considered central to the task of Natural Language Understanding, and there has been significant focus on the development of models that can perform well on the task~\cite{wang-etal-2018-glue}. This task of language inference has been independently extended to incorporate explanations~\cite{camburu-esnli} and figurative language~\cite{stowe-etal-2022-impli} (both detailed below). \newcite{https://doi.org/10.48550/arxiv.2205.12404} introduced \emph{FLUTE}, the Figurative Language Understanding and Textual Explanations dataset which brought together these two aspects. 

Previous shared tasks involving figurative language focused on the identification or representation of figurative knowledge: For example, FigLang2020~\cite{fig-lang-2020-figurative} and Task 6 of SemEval 2022~\cite{abu-farha-etal-2022-semeval} involved sarcasm detection, and Task 2 of SemEval 2022~\cite{tayyar-madabushi-etal-2022-semeval} involved the identification and representation of idioms. 

The generation of textual explanations necessitates the use of generative models such as BART \cite{lewis2020bart} or T5~\cite{DBLP:journals/corr/abs-1910-10683}. \newcite{narang2020wt5} introduce WT5, a sequence-to-sequence model that outputs natural-text explanations alongside its predictions and \citet{erliksson2021cross} found T5 to consistently outperform BART in explanation generation. 

Of specific relevance to our work are the IMPLI~\cite{stowe-etal-2022-impli} and  eSNLI~\cite{camburu-esnli} datasets. IMPLI links a figurative sentence, specifically idiomatic or metaphoric, to a literal counterpart, with the NLI relation being either entailment or non-entailment. \newcite{stowe-etal-2022-impli} show that idioms are difficult for models to handle, particularly in non-entailment relations. The eSNLI dataset \cite{camburu-esnli} is an explanation dataset for general NLI. It extends the Stanford Natural Language Inference dataset \cite{bowman2015large} with human-generated text explanations.

\emph{Hierarchical feature pipeline} based MTL architectures (\emph{HiFeatMTL}) use the outputs of one task as a feature in the next and are distinct from hierarchical \emph{signal} pipeline architectures wherein the outputs are used indirectly (e.g., their probabilities)~\cite{DBLP:journals/corr/abs-2109-09138}. HiFeatMTL has previously been used variously~\cite{10.1145/3292500.3330914,10.1609/aaai.v33i01.33016465,ijcai2020p536}, including, for example, to provide PoS and other syntactic information to relatedness prediction, the output of which is, in addition to the syntactic features, passed to an entailment task~\cite{hashimoto-etal-2017-joint} (see also the survey by ~\newcite{DBLP:journals/corr/abs-2109-09138}). 
To the best of our knowledge, this is the first work to use HiFeatMTL with, and to compare against, text-to-text models and their ability to transfer knowledge across tasks.

\section{Methods}
\label{oursetup}
We set out to answer the research questions in Section \ref{section:intro} by evaluating the effectiveness of SFT and HiFeatMTL on the transfer of task-specific knowledge from separate tasks, namely, explanations from eSNLI and figurative language from IMPLI. We use T5 for all our experiments as it has been shown to be effective in explanation generation~\cite{erliksson2021cross}. 
We run all our hyperparameter optimisation and model variations using T5-base (evaluated on a development split consisting of 10\% of the training data) before then transferring over the best performing settings to T5 large (trained on all of the training data) which is used to make predictions on the test set. While we find this method adequate in finding a good set of hyperparameters, the best setting for a smaller model need not necessarily be a good setting for larger models, especially given that some capabilities emerge only in larger models \cite{wei2022emergent}.

\subsection{Exploratory Experiments}
The first phase of our experiments was dedicated to using our development split to determining the best hyperparameters for T5, specifically the learning rate, and the number of beams, the two parameters that we found T5 to be extremely sensitive to. We do not experiment with prompt optimisation, but rather 
our prompts are based on what T5 was trained on (See listing \ref{lst:prompt}).

\lstset{breaklines=true,
        numbersep=5pt,
        xleftmargin=.25in,
        xrightmargin=.25in
} 
\begin{tiny}
\begin{lstlisting}[caption={Our default prompt used for T5.},captionpos=b,label={lst:prompt}]
Source_text:
    figurative hypothesis: <hypothesis>  premise: <premise>
target_text:
    <label> explanation: <explanation>
\end{lstlisting}
\end{tiny}

An additional consideration of this initial phase was whether it was more effective to independently perform the task of NLI before subsequently generating explanations. 
However, we find that incorporating the gold inference labels does not improve the quality of explanations generated.

\paragraph{Knowledge Transfer} To determine those forms of figurative language that T5 finds challenging and how effective knowledge transfer is, we test T5 fine-tuned just on FigLang2022, and sequentially on IMPLI followed by FigLang2022. The results of these experiments are presented in Table \ref{difficult-types}, which correspond to the observations made by \citet{stowe-etal-2022-impli} that idioms are particularly challenging for NLI models. Crucially, we find that the performance of the  model \emph{does} improve when first trained on IMPLI, 
thus establishing that knowledge transfer is possible in T5 through SFT. 

\begin{table}[!htbp]
\centering
\captionsetup{belowskip=-1.2\normalbaselineskip}
\resizebox{0.45\textwidth}{!}{
\begin{tabular}{lll}
\hline
\textbf{Type} & \textbf{FigLang} & \textbf{IMPLI $\rightarrow$ FigLang} \\\hline
Metaphor & 81.97 & \textbf{83.61 (+~2.0\%)} \\
\textbf{Simile} & \textbf{65.38} & \textbf{66.92 (+~1.5\%)}  \\
\textbf{Idioms} & \textbf{72.50} & \textbf{78.13 (+~6.0\%)}\\ 	
Creative Paraphrase & 98.36  & 98.36 \\
Sarcasm & 100 & 99.54 (-~0.5\%~) \\
\hline
\end{tabular}
}
\caption{T5 performance (acc) on the various labels of FigLang2022, before and after training on IMPLI.}
\label{difficult-types}
\end{table}

\vspace{2mm}

Importantly, we found that training for more epochs on the IMPLI dataset led to improved inference label accuracy but led to poorer explanations, which suggests knowledge transfer as oppsed to, for example, the advantage of additional training data. Since we were more interested in transferring figurative information from IMPLI, we optimise on Acc@0 (label accuracy) when training on IMPLI and Acc@60 (the evaluation metric relevant to the task) when training on the final FigLang dataset. 

\subsection{Experimental Setup}
\paragraph{Training Regime}In establishing the most effective method of knowledge transfer, we compare SFT with HiFeatMTL trained on: a) FigLang, b) eSNLI $\rightarrow$ FigLang, c) IMPLI $\rightarrow$ FigLang, d) eSNLI $\rightarrow$ IMPLI $\rightarrow$ FigLang, and e) IMPLI $\rightarrow$ eSNLI $\rightarrow$ FigLang. The training sets of both eSNLI and IMPLI are truncated to the same length as that of FigLang to ensure that the model does not over-fit on those other tasks.

\subsection{Sequential Fine-Tuning}
In SFT, we fine-tune the model on each of the relevant datasets in sequence. When training on the IMPLI dataset, which does not have associated explanations, we use the same prompt (Listing \ref{lst:prompt}) but with no associated explanation. The number of training epochs is established based on the change in loss on the development set and was found to be 3 for IMPLI and 10 for the other two datasets. 

\subsection{Multi-Task Learning}
\begin{figure}[!htb]
\centering
  \includegraphics[width=0.92\linewidth]{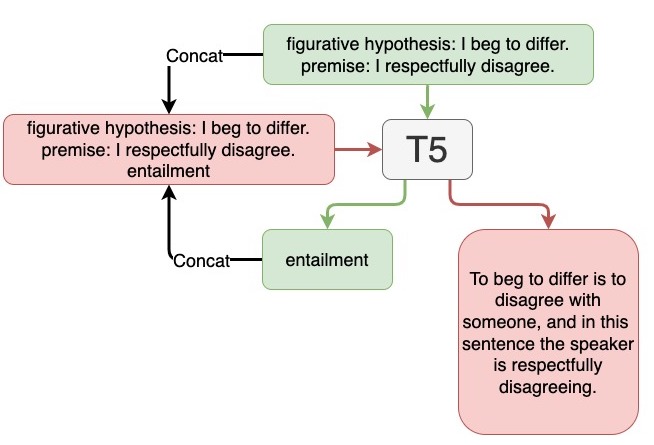}
  \caption{Our HiFeatMTL architecture. Note that we do not use GPT in our experiments, although it is possible to use GPT in place of T5.}
  \label{fig:e2e}
\end{figure}

We experiment with a \emph{hierarchical feature pipeline} for multi-task learning as the output inference label is likely to be important in generating the explanation. This involved creating an end-to-end model wherein, during the forward pass, T5 is used to predict the inference labels based on the hypothesis and the premise. This label, in addition to the hypothesis and premise are then used as input to T5 to generate an explanation. During the backward pass, the overall loss of the model is calculated as the weighted sum of the loss associated with each of the two steps above. Importantly, the weights of the T5 model used in the two steps are shared. Figure \ref{fig:e2e} provides an illustration.

As in the case of SFT, we fine-tune the model on each of the relevant datasets in sequence. When training on the FigLang dataset, we found it effective to train the model twice: fist with a higher weight to the loss associated with the inference (90\%) and a second time with a higher weight to the loss associated with explanations (also 90\%). Due to the summing of losses, we found that the model loss was not a good indicator of overfitting and instead determined the number of training epochs experimentally (10 for all datasets). 

\section{Results and Discussion}
\label{section:mainresults}

Table \ref{tab:st-results} shows the full shared task results from the CodaLab leaderboard\footnote{Our CodaLab submissions appear under the name  ``rachneet'': \url{https://codalab.lisn.upsaclay.fr/competitions/5908}} as of the competition's end date of 20 Aug, 2022. Our results (Team UKPChefs) are highlighted in bold. 

\begin{table}[!htbp] 
\centering
{\small
\begin{tabular}{@{} ccccc @{}}
\hline
\textbf{Rank} & \textbf{Team Name} &  \textbf{Acc@0} & \textbf{Acc@50} & \textbf{Acc@60} \\ \hline
\textbf{*1st} & \textbf{UKPChefs} & \textbf{0.925} &	\textbf{0.869} &	\textbf{0.633}  \\
*1st & TeamCoolDoge & 0.947 &	0.889 &	0.633 \\ 
2nd & vund & 0.936 &	0.865 &	0.607 \\ 
3rd & hoho5702 & 0.911 &	0.854 &	0.548  \\
4th  &  yklal95 & 0.847 &	0.779 &	0.517     \\ 
5th & tuhinnlp & 0.443 &	0.443 &	0.443   \\
6th & peratham.bkk & 0.590 &	0.203 &	0.033 \\
\hline
\multicolumn{2}{c}{\textit{Shared Task Baseline}} & \textit{0.817} &	\textit{0.748} &	\textit{0.483}  \\
\hline
\end{tabular}
\caption{Shared task results from all teams (ours -- UKPChefs -- in bold). Asterisks represent tied results.}
\label{tab:st-results}
}
\end{table}


%

\begin{table*}[!htbp] 
\centering
\captionsetup{belowskip=-0.8\normalbaselineskip}
{\footnotesize
\resizebox{0.85\textwidth}{!}{
\begin{tabular}{@{} cccccccccc @{}}
\hline
& \textbf{Dataset 1} & \textbf{Dataset 2} & \textbf{Dataset 3} & \multicolumn{2}{c}{\textbf{Acc@0}} & \multicolumn{2}{c}{\textbf{Acc@50}} & \multicolumn{2}{c}{\textbf{Acc@60}} \\ \hline
& & & & Dev & Test & Dev & Test & Dev & Test \\
\multirow{5}{*}{\rotatebox{90}{\small{SFT}}} 
& FigLang & - & - &84.99 & 93.27  & 78.49 & 87.80 & 56.18 & 61.74 \\ 
& eSNLI & FigLang & - & 86.06  & 92.67 & 80.74 & 87.20 & 57.77   & 63.27    \\ 
& IMPLI  & FigLang & - &86.59 & 93.20  &80.74 & 87.33 &56.97 & 60.93\\
& eSNLI & IMPLI & FigLang  &86.32 & 92.47 &80.08 & 86.87   &58.17 & 63.33       \\ 
& IMPLI  & eSNLI & FigLang  &84.99 &  92.73 &79.42& 87.33   &55.38 & 62.00 \\
\hline
\multirow{5}{*}{\rotatebox{90}{\small{HiFeatMTL}}} 
& FigLang & - & - & 91.24 & 94.67 & 82.07  & 86.54 & 55.11 & 55.13    \\ 
& eSNLI & FigLang & - & 91.50 & 94.14  & 82.07 & 86.40 & 55.91  & 53.80     \\ 
& IMPLI  & FigLang & - & 89.50 & N/A & 81.27  & N/A & 55.78 & N/A \\
& eSNLI & IMPLI & FigLang & 90.97 & 94.54  & 80.35 & 85.94 & 53.92 & 54.27      \\ 
& IMPLI  & eSNLI & FigLang & 89.37 & N/A & 80.34 & N/A  & 53.52  & N/A\\
\hline
\multicolumn{4}{c}{\textit{Shared Task Baseline}} & - & \textit{81.70} & - & \textit{74.80} & - & \textit{48.30} \\
\hline
\end{tabular}
}
\caption{Results of the SFT and HiFeatMTL models on the development and test splits of the FigLang2022 task. Experiments on the dev set were performed using T5-Base and those on the test set on T5-Large trained on the complete training set. Results marked N/A were not obtained due to the limits on the number of submissions.}
\label{main-results-table}
}
\end{table*}

The results of our experiments using SFT and HiFeatMTL are presented in Table \ref{main-results-table}. 
The results on the development set and those on the test set are not directly comparable: not only do we use different models, we also train on all the complete training data before evaluating on the test set. 
The drop in performance of the HiFeatMTL model on the test set on Acc@60, which consistently outperforming SFT on Acc@0 across both the development and the test sets is surprising. This seems to indicate that HiFeatMTL, while an effective way of boosting performance on the earlier of multiple dependent objectives, seems to be less effective on subsequent tasks (in this case, explanation generation). Additionally, HiFeatMTL also seems prone to overfitting, as the FigLang test set introduced novel idioms and similes previously unseen in the training set, into the test set. 

While the gain in accuracy when using the additional datasets could be due to the corresponding addition of training data, it should be noted that IMPLI does not have explanations and eSNLI contains no figurative language. As such, the improved scores indicate the transfer of figurative information from one task (IMPLI) and explanation generation capabilities from another (eSNLI).

As such, in addressing the research questions, our results indicate that: a) distinct task-specific knowledge (i.e. explanations or figurative language) can indeed be transferred from separate tasks so as to improve performance on a target task, and b) SFT seems to be a more effective way of transferring knowledge across tasks when we are concerned with the latter of a sequence of tasks (as in this case), while HiFeatMTL seems effective in boosting the performance of the first.

\section{Knowledge Transfer vs Bias}
\label{section:bias}

Recent works on NLI have shown that for some datasets, models are able to correctly predict the label using only the hypothesis, without considering the premise \cite{glockner-etal-2018-breaking, gururangan-etal-2018-annotation, mccoy-etal-2019-right}. This is caused by the model exploiting spurious correlations or patterns in the data, rather than acquiring task-relevant knowledge. As such, we wish to analyse if this is the case with our models: namely, whether our models employ figurative language knowledge from the hypothesis when predicting NLI labels. 

We perform the following experiments using T5 large on our validation set: we train only the hypothesis, only on the premise, and compare these results with a model trained on both (the standard training regime). The results 
(Table \ref{bias-results}) indicate that, while the model \emph{can} achieve reasonable accuracy while relying solely on the hypothesis, the significant improvement in accuracy (on both Acc@0 and Acc@60) when considering both the hypothesis and the premise indicates that, to a certain extent, the model is using knowledge of figurative language to predict the NLI labels and corresponding explanations.

\begin{table}[!htbp]
\centering
\captionsetup{belowskip=-1.2\normalbaselineskip}
\resizebox{0.40\textwidth}{!}{
\begin{tabular}{lccc}
\hline
\textbf{Setting} & \textbf{Acc@0} & \textbf{Acc@50}  & \textbf{Acc@60}  \\
\hline
Regular & 92.16 & 87.92 & 66.14 \\
Hyp-Only & 65.47 & 60.96 & 45.95 \\
Prem-Only & 56.31 & 47.81 & 33.74 \\
\hline
\end{tabular}
}
\caption{T5-large performance on the FigLang dataset with either the hypothesis or premise removed.}
\label{bias-results}
\end{table}

\section{Conclusions and Future work}
\label{section:conclusionsandfw}

In this work we set out to establish the possibility of effectively transferring knowledge across tasks in the context where we are interested in boosting the performance of two dependent tasks. As such, we evaluate the effectiveness of SFT and HiFeatMTL for transferring distinct task-specific knowledge from different tasks and find that both of these methods are good at achieving this: SFT on the last task and HiFeatMTL on the first. We find that using SFT to transfer information across tasks is, in this instance, so effective that we are \emph{ranked first} on the FigLang 2022 task. 

In extending this work, we intend to test these methods on a variety of sequentially dependent tasks as well as incorporating the use of more efficient MTL methods including AdapterFusion~\cite{pfeiffer-etal-2021-adapterfusion} and AdapterDrop~\cite{ruckle-etal-2021-adapterdrop}.


\section*{Acknowledgements}
This work was made possible through a research visit hosted by the UKP Lab\footnote{\url{https://www.informatik.tu-darmstadt.de/ukp/ukp_home/}} and funded by the Alan Turing Institute\footnote{\url{https://www.turing.ac.uk/}} through their Post-Doctoral Enrichment Award granted to HTM while at the University of Sheffield. In addition, this work was also partly supported by the UK EPSRC grant EP/T02450X/1, the European Regional Development Fund (ERDF), the Hessian State Chancellery – Hessian Minister of Digital Strategy and Development (reference 20005482, TexPrax), the State of Hesse in Germany (project 71574093, CDR-CAT), the German Federal Ministry of Education and Research and the Hessian Ministry of Higher Education, Research, Science and the Arts within their joint support of the National Research Center for Applied Cybersecurity ATHENE.

\section*{Limitations}
This work only deals with English, and since English makes up a majority of the training data for PLMs, performance may drop across other languages. Additionally, we only address figurative language within the context of the NLI task, and thus do not make broader claims about our model's ability to handle figurative language, to generate explanations or generalise across other generative models. This also extends to the comparisons between models that we present. 

\paragraph{Model Explanations} This work is involved in the generation of explanations associated with language inference predictions. Importantly, there is no guarantee (and very unlikely) that the generated explanations are indeed faithful to the process of predicting inference labels (also see~\newcite{jacovi-goldberg-2020-towards}). 

\paragraph{Carbon Footprint} All initial experiments are performed on smaller models and the best performing model architectures and parameters are transferred over to larger models to minimise the carbon footprint of our experiments. Despite this, the use of large language models does contribute to the climate crisis.

\bibliography{anthology,custom}
\bibliographystyle{acl_natbib}

\appendix

\end{document}